\documentclass[10pt,twocolumn,twoside]{IEEEtran}

\usepackage[switch]{lineno}
\usepackage[utf8]{inputenc} % allow utf-8 input
\usepackage[T1]{fontenc}    % use 8-bit T1 fonts
\usepackage{hyperref}       % hyperlinks
\usepackage{url}            % simple URL typesetting
\usepackage{booktabs}       % professional-quality tables
\usepackage{amsfonts}       % blackboard math symbols
\usepackage{amsmath}
\usepackage{microtype}      % microtypography
\usepackage{graphicx}
\graphicspath{ {images_for_paper/} }
\usepackage{multirow}
\usepackage{subcaption}
\usepackage{tabularx}

\newcommand{\bI}{\mathbf{I}}
\newcommand{\bJ}{\mathbf{J}}
\newcommand{\bA}{\mathbf{A}}
\newcommand{\bL}{\mathbf{L}}
\newcommand{\bW}{\mathbf{W}}
\newcommand{\bU}{\mathbf{U}}
\newcommand{\bT}{\mathbf{T}}
\newcommand{\bx}{\mathbf{x}}
\newcommand{\by}{\mathbf{y}}
\newcommand{\bt}{\mathbf{t}}
\newcommand{\bSigma}{\boldsymbol{\Sigma}}

\newcommand{\bmu}{\boldsymbol{\mu}}
\DeclareMathOperator*{\argmin}{arg\,min}

\newcommand{\hazy}[1]{\includegraphics[width=0.14\textwidth]{hazy/#1}}
\newcommand{\DCP}[1]{\includegraphics[width=0.14\textwidth]{DCP/#1}}
\newcommand{\BCCR}[1]{\includegraphics[width=0.14\textwidth]{BCCR/#1}}
\newcommand{\NLD}[1]{\includegraphics[width=0.14\textwidth]{NLD/#1}}
\newcommand{\CAP}[1]{\includegraphics[width=0.14\textwidth]{CAP/#1}}
\newcommand{\MSCNN}[1]{\includegraphics[width=0.14\textwidth]{MSCNN/#1}}
\newcommand{\AODNet}[1]{\includegraphics[width=0.14\textwidth]{AODNet/#1}}
\newcommand{\DehazeNet}[1]{\includegraphics[width=0.14\textwidth]{DehazeNet/#1}}
\newcommand{\GFN}[1]{\includegraphics[width=0.14\textwidth]{GFN/#1}}
\newcommand{\ours}[1]{\includegraphics[width=0.14\textwidth]{Ours/#1}}
\newcommand{\clean}[1]{\includegraphics[width=0.14\textwidth]{clean/#1}}

\newcommand{\hazyq}[1]{\includegraphics[width=0.1\textwidth]{hazy/#1}}
\newcommand{\DCPq}[1]{\includegraphics[width=0.1\textwidth]{DCP/#1}}

\newcommand{\NLDq}[1]{\includegraphics[width=0.1\textwidth]{NLD/#1}}
\newcommand{\CAPq}[1]{\includegraphics[width=0.1\textwidth]{CAP/#1}}
\newcommand{\MSCNNq}[1]{\includegraphics[width=0.1\textwidth]{MSCNN/#1}}
\newcommand{\AODNetq}[1]{\includegraphics[width=0.1\textwidth]{AODNet/#1}}
\newcommand{\DehazeNetq}[1]{\includegraphics[width=0.1\textwidth]{DehazeNet/#1}}
\newcommand{\GFNq}[1]{\includegraphics[width=0.1\textwidth]{GFN/#1}}
\newcommand{\oursq}[1]{\includegraphics[width=0.1\textwidth]{Ours/#1}}
\newcommand{\oursp}[1]{\includegraphics[width=0.1\textwidth]{Ours_30/#1}}

\newcommand{\ITSorig}[1]{\includegraphics[width=0.16\textwidth]{ITS_orig/#1}}
\newcommand{\ITShazy}[1]{\includegraphics[width=0.16\textwidth]{hazy/#1}}
\newcommand{\ITStrans}[1]{\includegraphics[width=0.16\textwidth]{gt_t/#1}}

\begin{document}
%\linenumbers

\title{Unsupervised Single Image Dehazing Using Dark Channel Prior Loss\\}

\author{Alona~Golts,
        Daniel~Freedman,
        and~Michael~Elad,~\IEEEmembership{Fellow}% <-this % stops a space
\thanks{A. Golts and M.Elad are from the Department
of Computer Science, Technion Institute of Technology, Technion City, Haifa 32000, Israel,
corresponding e-mails: salonaz@cs.technion.ac.il, elad@cs.technion.ac.il.
D. Freedman is from Google Research, Haifa, Israel, email: danielfreedman@google.com.}% <-this % stops a space
}

\markboth{IEEE TRANSACTIONS ON IMAGE PROCESSING,~Vol.~X, No.~Y, October~2018}%
{Shell \MakeLowercase{\textit{et al.}}: Bare Demo of IEEEtran.cls for IEEE Journals}

\maketitle

\begin{abstract}
Single image dehazing is a critical stage in many modern-day autonomous vision applications. Early prior-based methods often involved a time-consuming minimization of a hand-crafted energy function. Recent learning-based approaches utilize the representational power of deep neural networks (DNNs) to learn the underlying transformation between hazy and clear images. Due to inherent limitations in collecting matching clear and hazy images, these methods resort to training on synthetic data; constructed from indoor images and corresponding depth information. This may result in a possible domain shift when treating outdoor scenes. We propose a completely unsupervised method of training via minimization of the well-known, Dark Channel Prior (DCP) energy function. Instead of feeding the network with synthetic data, we solely use real-world outdoor images and tune the network's parameters by directly minimizing the DCP. Although our ``Deep DCP'' technique can be regarded as a fast approximator of DCP, it actually improves its results significantly. This suggests an additional regularization obtained via the network and learning process. Experiments show that our method performs on par with large-scale supervised methods.

\end{abstract}


\section{Introduction}

Haze is an atmospheric phenomenon where small particles, called aerosols, obstruct the clarity of an outdoor scene and lead to poor contrast and loss of detail. The existence of haze affects an image in two aspects. It attenuates the scene radiance with correspondence to an object's distance from the camera. Moreover, it introduces an additional ambient light component, called the \textit{airlight}, which causes a ``veiling effect'' over the clear image. The formation of a hazy image is often described as a linear per-pixel combination of the clear scene radiance and the airlight; the effect of each component is controlled by the transmission map. To recover the scene radiance image, one has to solve a system of $3N$ linear equations with $4N+3$ unknowns (where $N$ is the number of image pixels).

In order to handle the under-constrained haze creation model, many researchers suggested hand-crafted image priors, shedding additional light on the behaviour of hazy versus clean images \cite{fattal_dehazing, contrast_tan, DCP, tarel_dehazing, color_lines, CAP, NLD, BCCR, hue_dehazing}. These prior-based methods often formulate the problem of dehazing as an energy minimization task, where obtaining the solution of each image is called ``inference'', requiring a non-trivial optimization scheme.
With the increasing importance of image dehazing as an initial pre-processing stage in many computer-vision tasks (e.g., object detection, autonomous car navigation), large-scale learning-based techniques have been deployed to solve it \cite{aodnet,dehazenet,mscnn,GFN, VGG_dehazing}. These methods, however, require thousands of input and output examples.

Since clean and hazy images of the exact same scene and lighting conditions are hard to obtain, learning-based methods commonly resort to synthetic dataset creation. Given a clean image and a corresponding depth map, one can calculate the transmission map and use the haze creation model to obtain hazy images with varying amounts of haze and airlight components. These pairs of hazy and clear images are later fed as inputs and labels in a supervised training of a DNN. 
Outdoor depth information, however, is incredibly imprecise. For instance, the depth information of the outdoor Make3D \cite{make3d} and KITTI \cite{kitti} datasets suffers from over 4 meters of average root-Mean-Square-Error (rMSE), while the rMSE of the indoor NYU2 \cite{NYU2} is only 0.5. Consequently, large-scale methods either use the more reliable indoor depth information \cite{mscnn,aodnet,GFN,VGG_dehazing}, or draw the depth map at random \cite{CAP,dehazenet}. Either of these practices creates a domain shift when addressing real-world outdoor images. 

We propose to leverage the representational power of DNNs, but instead of feeding them with inaccurate synthetic pairs of hazy and clean images, we train them in an unsupervised fashion using real-world hazy images only. We optimize the network's weights by minimizing an unsupervised loss function, essentially the Dark Channel Prior (DCP) \cite{DCP} energy function. Our network can be regarded as a fast feed-forward approximator of the DCP. However, by stopping the optimization early, we get a significant boost in results over the classic DCP. This implies an added regularization, stemming from the network architecture and learning process. Our network, based on the Context Aggregation Network (CAN) architecture \cite{CAN}, is trained end-to-end from scratch without relying on any external data apart from raw hazy images. It provides the predicted transmission maps as output, from which the dehazed image can be easily reconstructed. We perform a comprehensive quantitative evaluation of our method and present state-of-the-art results on \textit{SOTS-outdoor} in the recently released RESIDE dataset \cite{reside}. We show qualitative results on real-world images, demonstrating that the additional regularization provided by the network reduces common artifacts of prior-based methods, such as over-saturation and high-contrast.

Our ``Deep-DCP'' method offers the following contributions: 
\begin{enumerate}
\item It provides state-of-the-art results in outdoor single image dehazing, outperforming both prior-based and fully-supervised DNN methods.
\item It achieves an impressive $\sim 6.5$dB boost in outdoor PSNR over classical DCP, validating an effective regularization.
\item It treats the sky successfully where DCP typically fails.
\item It is the first to perform unsupervised training in single image dehazing, reducing the need in synthetic data.
\item It does not require an explicit optimization for each image as DCP, but rather learns the underlying transformation during training, requiring a fast forward-pass during test.
\item It offers a generic methodology of unsupervised training with energy functions and can be applied to any successful energy function.
\end{enumerate}

The remainder of this paper is structured as follows: Section \ref{s:related_work} provides a survey of previous prior-based and data-driven approaches for dehazing; Section \ref{our_method} describes the DCP, its use as a loss function and our CAN-based architecture; Section \ref{experiments} provides quantitative, qualitative and runtime experimental results; Section \ref{discussion} includes discussions and further analysis; finally, Section \ref{conclusions} concludes this work.

\section{Related Work}\label{s:related_work}

\subsection{Prior-Based Approaches}
Early attempts at image dehazing have incorporated several images of the same scene, taken at different bad weather conditions \cite{multiple_image_dehazing}, or using different polarization filters  \cite{multiple_polar_dehazing}. Kopf \textit{et al.} \cite{dehazing_with_depth} later performed dehazing of outdoor images by utilizing existing geo-referenced terrain and urban models including depth, texture and GIS data.

In \cite{contrast_tan}, Tan \textit{et al.} unveiled the haze from a single image by maximizing the local contrast of each patch in the image using a Markov Random Field (MRF) framework. In \cite{fattal_dehazing}, Fattal \textit{et al.} suggested utilizing the lack of correlation between the transmission and shading in a localized set of pixels, as a prior to resolve the ambiguity between the scene albedo and the airlight.
Tarel \textit{et al.} \cite{tarel_dehazing} provided a fast calculation of the ``atmosperic veil'' using a series of edge-preserving linear filter operations. In \cite{nishino_dehazing}, Nishino \textit{et al.} exploited the statistical independence between the scene albedo and depth and factorized both quantities into an MRF-based energy function. 

In \cite{DCP}, He \textit{et al.} proposed the now widely used DCP and demonstrated that in clear images the darkest pixel in an image patch is close to zero (this, however, does not hold in sky-regions).
Using this and the assumption that the transmission map within a small image patch is constant, a coarse map can be easily derived. They further suggested a computationally costly soft matting operation for smoothing out the transmission and reconstructing the final dehazed image. Follow-up works have improved both the quality and efficiency of DCP. Specifically, in \cite{BCCR}, the authors proposed a general boundary constraint for the transmission map for which the DCP is a special case.

Several color-based priors have been suggested as well for boosting dehazing performance \cite{color_lines,color_ellipsoid,NLD}. In \cite{color_lines}, Fattal used the ``color-lines'' assumption, stating that pixels in small image patches have a one-dimensional distribution in RGB-space \cite{color_lines_assumption}. The offset of these straight lines from the origin in hazy images allow to estimate the transmission map. Berman \textit{et al.} proposed a global approach, called non-local dehazing (NLD) \cite{NLD}. They observed that a haze-free image contains only several hundreds of distinct colors, clustered as points in RGB-space. In the presence of haze, these color clusters form a ``haze-line'' where the position of a certain pixel along the line corresponds to its initial radiance color and distance from the camera.

While prior-based methods reveal fine image details, they often suffer from increased saturation and contrast, unrealistic colors and difficulty in handling sky regions. This is due in part to assumptions not suited for all hazy image patches. In addition, each image requires a separate non-trivial optimization and solution which can be prohibitive for real-time applications.

\subsection{Data-Driven Approaches}

In \cite{CAP}, a Color Attenuation Prior (CAP) is suggested, mixing hand-crafted observations with a data-driven approach. CAP assumes that the image depth, the amount of haze and the difference between the brightness and saturation are linearly correlated. To find the exact correlation, the authors opt for supervised regression between synthesized hazy patches and their corresponding depth maps. This results in fast inference at test time.

One of the first works to propose single image dehazing using CNNs is \cite{mscnn}. The method, called MSCNN, is trained by feeding a two-stage network with pairs of hazy images and corresponding transmission maps. In DehazeNet \cite{dehazenet}, Cai \textit{et al.} create a novel CNN architecture
(featuring maxout and BReLU layers), inspired by popular prior-based methods \cite{DCP,contrast_tan,CAP,hue_dehazing}. AOD-Net \cite{aodnet} in turn, proposes a joint estimation of both the transmission map and the airlight via a unified representation. Using this representation, one can easily reconstruct the scene radiance directly in an end-to-end forward-pass computation. This helps reduce errors accumulated in the separate calculation of the two quantities.

In the recent Gated Fusion Network (GFN) \cite{GFN}, a dehazed image is produced as a fusion of the white balance, contrast enhanced and gamma corrected images (all derived from the hazy image). The network outputs three confidence maps which determine the effect of each component. To combat halo effects of a single scale encoder-decoder structure, a multi-scale architecture is used where a coarse output is first produced, then added as input to a finer scale network. This method provides impressive results on RESIDE's \textit{SOTS-indoor}, but quadruples the size of the input during training and test, making evaluation inefficient in terms of memory. Finally, in a recent work reported in \cite{VGG_dehazing}, the authors utilize the pre-trained VGG \cite{VGG} network as encoder and only train the symmetric decoder via a combination of MSE and perceptual loss.

While learning-based methods achieve impressive results, they are trained in a supervised way, relying on synthetic datasets. Some methods use more accurate \textit{indoor} depth information to create labelled inputs \cite{mscnn,aodnet,GFN,VGG_dehazing}. This practice, however, directs increasing research effort to optimizing indoor performance, while the predominant need for dehazing is actually outdoors. 
Other methods use real-world \textit{outdoor} images, but compromise the accuracy of the depth information. For example, \cite{CAP} draws each pixel in the depth map at random from a $(0,1)$ uniform distribution and \cite{dehazenet} enforces an additional constraint of constant depth within $16 \times 16$ patches. 
These assumptions result in block and halo artifacts in the reconstructed image and require additional post-processing.

\section{Our Method}\label{our_method}
In the following we will describe our method for single image dehazing, including the driving force of our unsupervised loss function, the Dark Channel Prior \cite{DCP}, its implementation as a loss function for training a CNN and the architecture we choose for the task at hand.

\subsection{Haze Model}\label{haze_formation}

The popular haze formation model in \cite{formation_model} is given as:
\begin{equation}\label{eq:haze_model}
\begin{split}
	\bI(\bx) &= t(\bx)\bJ(\bx) + (1-t(\bx))\bA,\\
	t(\bx) &= e^{-\beta d(\bx)}. \\
\end{split}
\end{equation}
According to the above, the observed hazy image, $\bI(\bx) \in \mathbb{R}^{N \times 3}$, is a convex linear combination of the haze-free scene radiance, $\bJ(\bx)$, and the atmospheric light component, $\bA$, called the \textit{airlight}; usually represented as a constant 3-vector in RGB-space, $\bA = \left(A^r,A^g,A^b\right)$. The transmission map coefficients, $t(\bx) \in \mathbb{R}^N$ control the relative force of each component, in each pixel in the image, $\bx \in \mathbb{R}^N$. The transmission is a function of the depth, $d(\bx)$, of the scene from the observer. Our goal in single-image-dehazing is to obtain the haze-free scene radiance, $\bJ(\bx)$. To do so, however, one needs to solve a set of $3N$ equations (only $\bI(\bx)$ is given), with $4N+3$ unknowns ($\bJ(\bx), t(\bx),\bA$). Thus, additional prior knowledge of the images in question is needed.

\subsection{Dark Channel Prior}\label{dark_channel}

The dark channel prior is an image statistical property, indicating that in small patches of haze-free outdoor images, the darkest pixel across all color channels is very dark, and close to zero. The ``dark channel'' of the image is defined as
\begin{equation}
	\bJ^{dark}(\bx) = \underset{c \in \{r,g,b\}}{\min} (\underset{\by \in \Omega(\bx)}{\min}  \left(\bJ^c(\by)\right)),
\end{equation}
where $\Omega(\bx)$ is a small patch, centered around $\bx$. This observation is contributed by three factors which appear in outdoor images: (1) shadows -- induced by cars, buildings and trees; (2) colorful objects -- where one color channel is dominant, and the others are close to zero, e.g., red flowers, green leaves, blue sea; and (3) naturally dark objects -- such as tree trunks and rocks.

Assuming that $\bA$ is known and the transmission within a small image patch, denoted as $\tilde{t}(\bx)$, is constant, one can apply a minimum operation across channels and pixels in the haze formation equation in (\ref{eq:haze_model}) (effectively zeroing $\bJ^c(\by)$) and get a prediction for the transmittance \cite{DCP}:
\begin{equation}
	\tilde{t}(\bx) = 1 - \omega \cdot \underset{c}{\min} \left( \underset{\by \in \Omega(\bx)}{\min} \left( \frac{\bI^c(\by)}{A^c} \right) \right),
\end{equation}
where $\omega=0.95$ leaves a small amount of haze for natural-looking results. In sky regions although the dark channel does not always hold, it is assumed that $\bI/\bA \rightarrow 1$, thus $\tilde{t}(\bx) \rightarrow 0$. The resulting coarse transmission map requires an additional step of refinement.

\subsection{Soft Matting}\label{soft_matting}

The haze formation model in (\ref{eq:haze_model}) is very similar to the composition model in image matting \cite{soft_matting}, where an output image is a convex linear combination of foreground and background images; controlled by the alpha matte, $\alpha$. If one replaces the $\alpha$-matte with the coarse transmission map, $\tilde{t}(\bx)$, the following energy function suggested in \cite{soft_matting} can be used to acquire the refined map, $t(\bx)$:
\begin{equation}\label{eq:soft_matting}
	E(\bt,\tilde{\bt}) = \bt^T \bL \bt + \lambda (\bt - \tilde{\bt})^T (\bt - \tilde{\bt}),
\end{equation}
where the first term promotes successful image matting, and the second, fidelity to the dark channel solution. The parameter $\lambda$, controlling the force between the two, is set to $\lambda=10^{-4}$ \cite{DCP}. The matrix $\bL$ is a Laplacian-like matrix, dedicated to image matting and given by \cite{soft_matting}:
\begin{equation}\label{eq:matting_weights}
\begin{split}
  \bL_{ij} & = \sum_{n | (i,j) \in p_n} (\delta_{ij} - w_{ij}^n), \quad  \forall i,j=1...N\\
   w_{ij}^n & = \scriptstyle \frac{1}{|p_n|} \left(1 + (\bI_i - \bmu_n)^T(\bSigma_n + \frac{\varepsilon}{|p_n|}\bU_3)^{-1}(\bI_j - \bmu_n)\right), \\
\end{split}
\end{equation}
where $i,j$ are two pixels within a small patch $p_n$ around pixel $n$; $|p_n|$ is the size of the patch and equal to $3 \times 3=9$ as suggested in \cite{soft_matting}; $\bmu_n \in \mathbb{R}^3$ and $\bSigma_n \in \mathbb{R}^{3 \times 3}$ are the mean and covariance of the patch; $\bU_3$ is the identitly matrix; and $\varepsilon$ is a smoothing parameter set to $\varepsilon=10^{-6}$ \cite{soft_matting}.

\begin{figure*}[t!]
\centering
\includegraphics[width=0.87\textwidth]{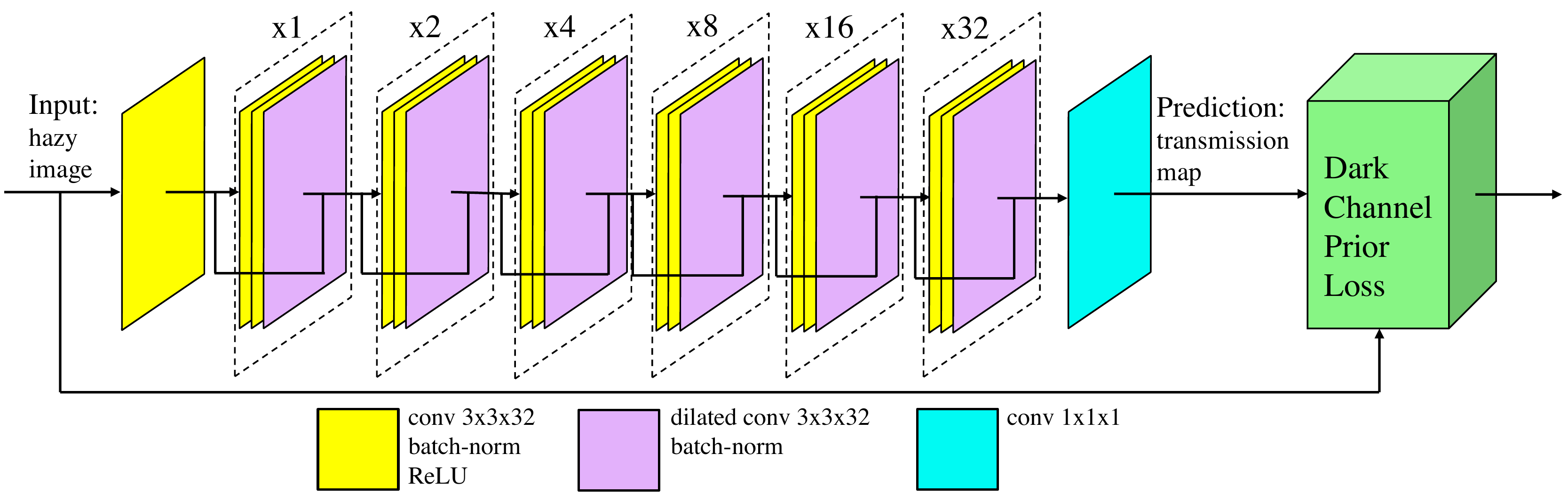}
\caption{System architecture. Our fully-convolutional network receives real-world hazy images. Apart from the input and output layers, our network is a cascade of dilated residual blocks (dilation written above each block), which gradually increase the receptive field. The network's predicted transmission and the input image, are fed to the unsupervised, DCP loss.}
\label{fig:architecture}
\end{figure*}
\vspace{-1mm}

\subsection{Implementation as a Loss Function}

We rewrite the energy function in equation (\ref{eq:soft_matting}) in a tensor-friendly format by using a known decomposition of Laplacian matrices via their weights, given in (\ref{eq:matting_weights}). Rephrasing the first term in (\ref{eq:soft_matting}) in terms of weights, we have that
\begin{equation}
E_1(\bt,\tilde{\bt}) = \bt ^T \bL \bt = \sum_{n=1}^N \sum_{i=1}^9 \sum_{j=1}^9 w_{ij}^n (t_i - t_j)^2,
\end{equation}
where we sum over all overlapping patches around $N$ pixels in the resulting transmission map, $\bt$, as well as over all possible combinations of pixel pairs, $i,j$, in a given $3 \times 3$ patch. The maximum number of combinations is $(3^2)\cdot(3^2)=81$. We can vectorize this term, along with the data fidelity term
\begin{equation}\label{eq:loss_function}
E(\bt,\tilde{\bt}) = \sum_{n=1}^N \sum_{k=1}^K \bW \odot ({\bT}_I - {\bT}_J)^2
    + \lambda \sum_{n=1}^N (\bt-\tilde{\bt})^2,
\end{equation}
where $\odot$ denotes elementwise multiplication; $k \in [1..81]$ indexes all possible pairs of pixels in a $3 \times 3$ patch, and $\bW \in \mathbb{R}^{N \times 81}$ is the vectorized version of the weights. ${\bT}_I,{\bT}_J \in \mathbb{R}^{N \times 81}$ are repetitions of the output transmission map. The first representing the transmission patches ($9$ pixels in total) arranged in $I \rightarrow (1,..,1,2,...,2,...,9,...,9) \in \mathbb{R}^{81}$, and the second arranged in $J \rightarrow (1,2,...,9,1,2,...,9,...,1,2,...9) \in \mathbb{R}^{81}$. 

\begin{figure}[t!]
\centering
\includegraphics[width=0.45\textwidth]{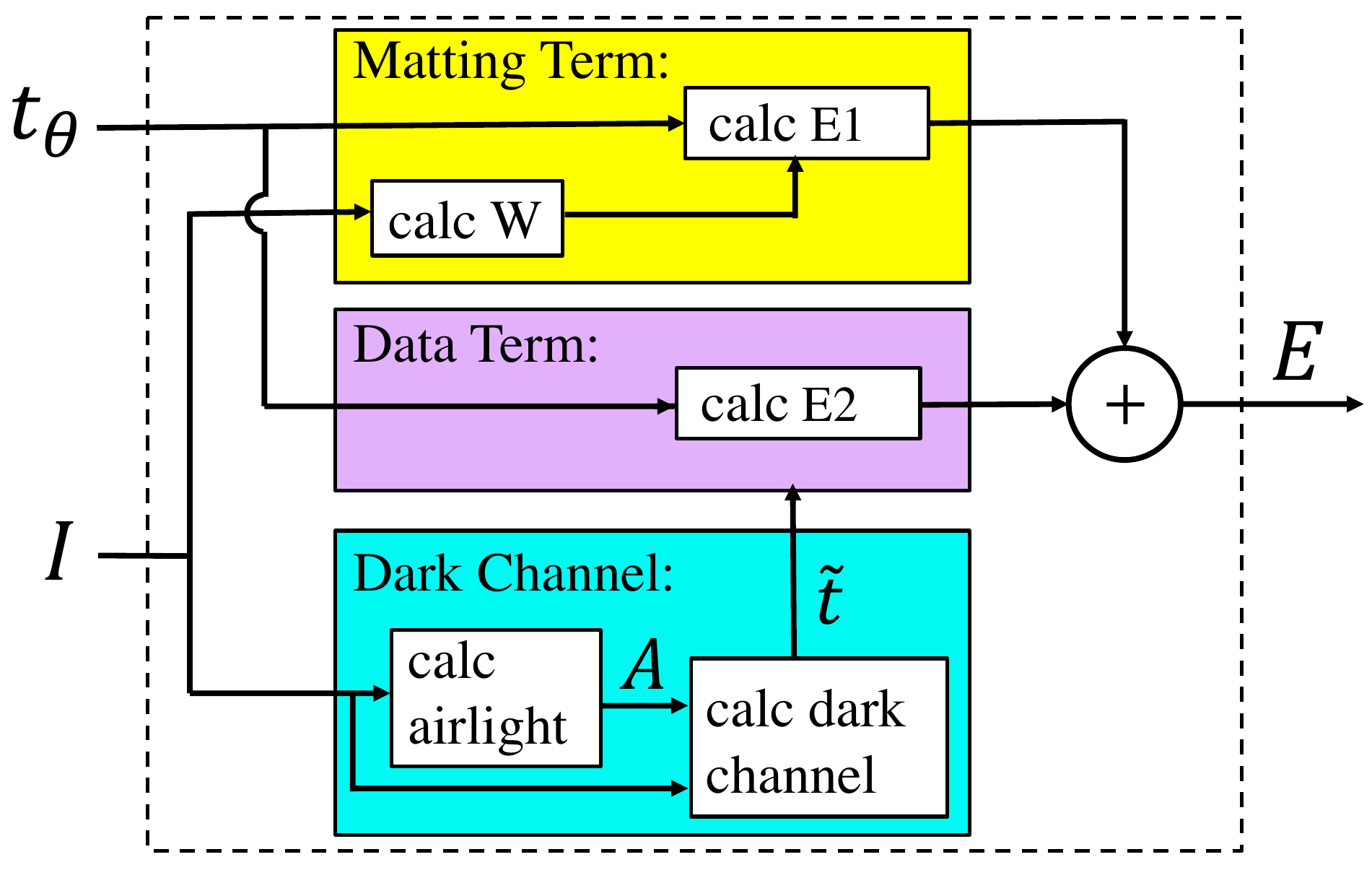}
\caption{Our loss module, which receives the prediction of the network, $\bt_{\theta}$, along with the hazy image, $\bI$, and outputs the value of the DCP \cite{DCP} energy loss.}
\label{fig:dehazing_loss}
\end{figure}

Above is the loss function with which we train our network, whose predicted transmission map is parametrized by $\bt_{\theta}$. We tune the parameters, $\theta$, by minimizing the loss function in (\ref{eq:loss_function}) over the training set of hazy images, $\{\bI_m\}_{m=1}^M$:
\begin{equation}
\theta^* = \underset{\theta}{\argmin} \left[\frac{1}{M} \sum_{m=1}^{M} E(\bt_{\theta},\tilde{\bt}(\bI_m))\right],
\end{equation}
where $M$ is the number of images. Note that we do not use the ``labels'', i.e., the clear images, at any point, only the original hazy ones. A schematic diagram of the inputs and outputs of our loss module is given in Figure. \ref{fig:dehazing_loss}.

\subsection{Computing the Scene Radiance}
Once the network has finished training, the transmission map, $t_{\theta}(\bx)$, of a new hazy image can be obtained by a forward-pass operation. This is used to recover the scene radiance via the haze formation model in (\ref{eq:haze_model}):
\begin{equation} \label{eq:reconstruction}
\bJ(\bx) = \frac{\bI(\bx) - \bA}{\max (t_{\theta}(\bx),t_0)} + \bA,
\end{equation}
where $t_0$, which discourages division by numbers close to zero, is set to $t_0=0.1$ as suggested in  \cite{DCP}. In order to recover the missing airlight component, $\bA$, we follow the method suggested in \cite{DCP}: we first pick the $0.1\%$ brightest pixels in the dark channel of the hazy image. Then, out of these locations we pick the brightest pixel in the hazy image, $\bI$. That is the final chosen atmospheric light, $\bA$.

\subsection{Architecture}
Our fully-convolutional, ``Dilated Residual Network'', shown in Figure \ref{fig:architecture}, is inspired by the Context Aggregation Network (CAN) \cite{CAN}, which has shown impressive results in dense-output applications. Similarly to CAN, we keep the resolution of all layers intact and identical to that of the input and output. In order to get an accurate prediction we avoid pooling and upsampling, and instead increase the receptive field via dilated convolutions with exponentially increasing dilation factors. Contrary to \cite{CAN}, between each dilated convolution we add another two regular convolution layers to create a richer nonlinear representation.

Our network is thus built as a cascade of $6$ dilated residual blocks; each made up of two regular convolutions, followed by a single dilated convolution. The dilation factors increase by a power of two from one block to the next. The filter size and width of all convolution layers (apart from the output) is $3 \times3 \times 32$. All regular convolutions are followed by batch normalization \cite{batch_norm} and ReLU nonlinearity \cite{alexnet}, and all dilated ones are followed by batch-norm only. The final layer is a linear transformation to the output dimension of the transmission map $1 \times 1 \times 1$.
To improve gradient flow and propagate finer details to the output, we incorporate additional Resnet-style \cite{resnet} skip connections between the input and output of each block. The skip connection is a simple addition of the input to the output of each block.

\section{Experimental Results}\label{experiments}

\begin{table*}[t!]
\begin{center}
\resizebox{.96\textwidth}{!}{
\begin{tabular}{l c c c c c c c c c}
\hline
\hline
 & DCP \cite{DCP} & BCCR \cite{BCCR} & NLD \cite{NLD} & CAP \cite{CAP} & MSCNN \cite{mscnn} & DehazeNet \cite{dehazenet} & AOD-Net \cite{aodnet} & GFN \cite{GFN} & Ours  \\
\hline
\textit{HSTS} & 17.22/0.798 & 15.09/0.738 & 17.62/0.792 & 21.54/0.867 & 18.29/0.841 & \textbf{24.49}/0.915 & 21.58/0.922 & 22.94/0.874 & 24.44/\textbf{0.933}  \\
\textit{SOTS-outdoor} & 17.56/0.822 & 15.49/0.781 & 18.07/0.802 & 22.30/0.914 & 19.56/0.863 & 22.72/0.858 & 21.34/0.924 & 21.49/0.838 & \textbf{24.08}/\textbf{0.933} \\
\textit{SOTS-indoor} & 20.15/0.872 & 16.88/0.791 & 17.29/0.749 & 19.05/0.836 & 17.11/0.805 & 21.14/0.847 & 19.38/0.849 & \textbf{22.32}/\textbf{0.880} & 19.25/0.832 \\
\hline
\hline
\end{tabular}
}
\end{center}
\caption{Quantitative PSNR/SSIM results of our approach (higher is better). For both SOTS-outdoor and HSTS we report the result of epoch 27, whereas in SOTS-indoor we report the result of epoch 30.}\label{tbl:results}
\end{table*}
\vspace{-1mm}

\subsection{Dataset}
In order to train and evaluate the performance of our network, we use the recent large-scale RESIDE (REalistic Single Image DEhazing) dataset \cite{reside}. RESIDE's training set, called \textit{``ITS''}, includes $13,990$ synthetic indoor images, created from the NYU2 \cite{NYU2} and Middlebury stereo datasets \cite{middlebury}. The test set includes both indoor and outdoor sections, called \textit{``SOTS-indoor''} and \textit{``SOTS-outdoor''}\footnote{Although in the latest published paper of RESIDE, SOTS-outdoor is not officially featured, the selection of $500$ specific outdoor images are still available (as well as in earlier Arxiv versions) in RESIDE's website in: \url{https://www.dropbox.com/s/y6jupfvitv0dx5w/SOTS.zip?dl=0&file_subpath=\%2FSOTS}},
each containing 500 synthetic images. A smaller test set of $20$ outdoor images, called \textit{``HSTS''}, is also suggested. HSTS has a mix of $10$ synthetic images (where ground truth is known) and $10$ real-world images. All synthetic hazy images are created by first collecting ground-truth clean images with their corresponding depth maps and applying the haze formation model with different configurations of the $\bA,\beta$ parameters in (\ref{eq:haze_model}). The beta version of RESIDE provides an additional collection of $4,322$ real-world images, mined from the web, called \textit{``RTTS''}. Instead of using the synthetic indoor database of ITS (or its variations based on NYU2 and Middlebury), as in \cite{GFN,mscnn,aodnet,VGG_dehazing}, we train our network on the real-world images of RTTS. For the evaluation of PSNR (Peak Signal to Noise Ratio) and SSIM (Structural Similarity) criteria during training, we use as validation a subset of $500$ images from RESIDE beta's \textit{``OTS''} synthetic outdoor training set. The $500$ images are selected at random from Part-I of OTS, where we make sure that no images from the SOTS-outdoor and HSTS test sets are selected for validation.

\subsection{Implementation Details}
To enrich the RTTS training set, we perform data augmentation. The first augmentation is simply resizing the original hazy images to size $128 \times 128$ using bilinear interpolation. The second, third and fourth augmentations are performed randomly. Each image can be flipped horizontally or kept as is; randomly cropped to $256 \times 256$ or $512 \times 512$, and rotated at $0,45,90$, or $135$ degrees. If rotated, only the valid center of the image is taken. All augmented images are then resized to $128 \times 128$. The final number of training images is therefore: $4322 \times 4=17,288$.

The parameters of our loss function are taken exactly (no additional tuning) as suggested in \cite{DCP,soft_matting}: $\lambda=10^{-4}$, $\omega=0.95$, $t_0=0.1$, $\varepsilon=10^{-6}$, DCP patch size: $15 \times 15$, and soft matting patch size: $3 \times 3$. We use the Adam optimizer \cite{adam} with batch size of $24$; initial learning rate of $l_r=3 \cdot 10^{-4}$, and exponential decay with factor $0.96$ every $3$ epochs. The network weights are initialized using random initialization with zero mean and variance of $0.1$. Our method is implemented in TensorFlow on a GTX Titan-X Nvidia GPU. Training time to get the optimal solution (about $30$ epochs) takes $8$ hours. For outdoor results we stop the training at epoch $27$, whereas for SOTS-indoor, we keep training until we reach $30$ epochs. Our stopping criterion is explained further in section \ref{stopping_criterion}.

\begin{figure*}
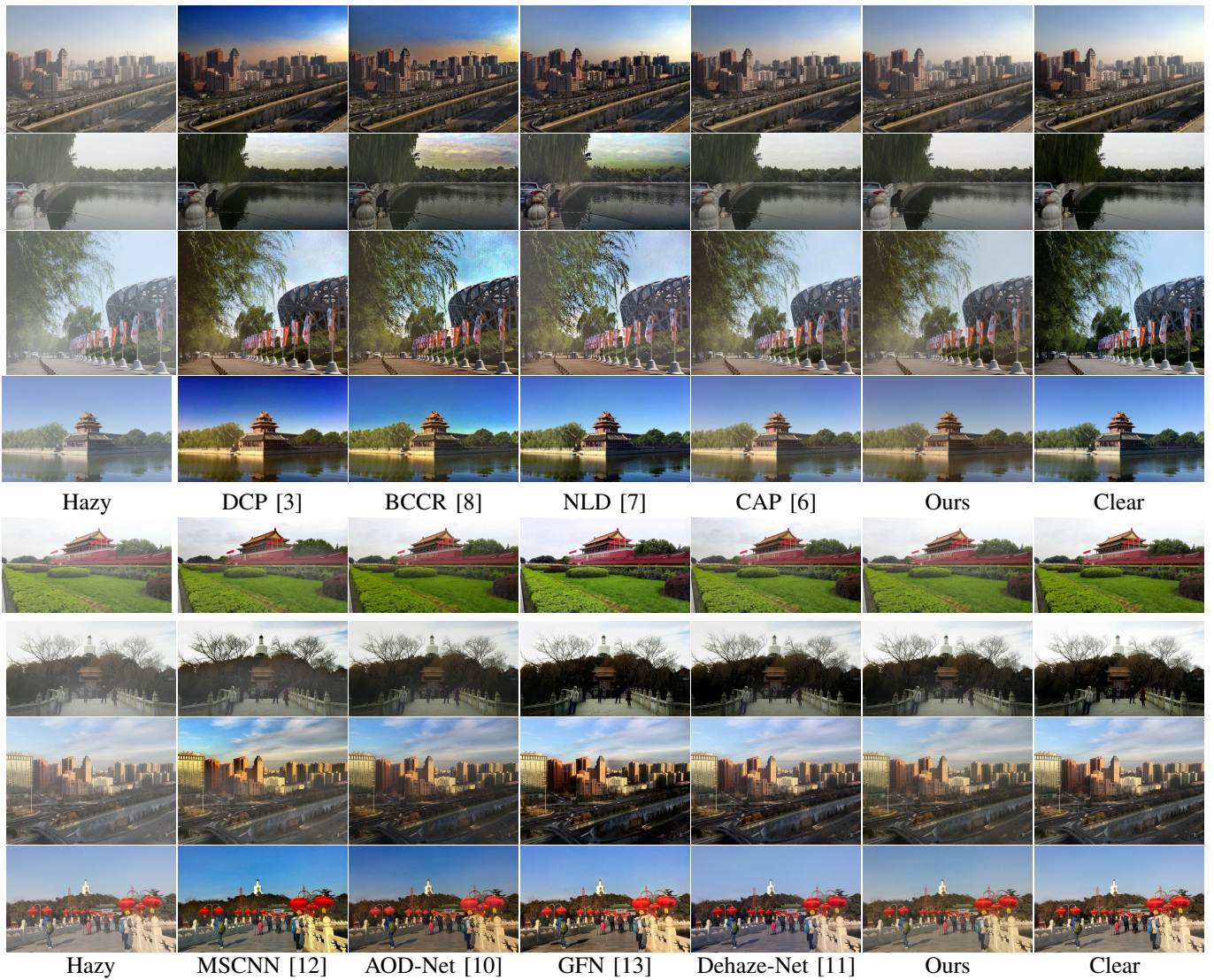

    \begin{center}
    \setlength{\tabcolsep}{0.5pt}
    \begin{tabular}{cccccccccc}

	\vspace{-1mm}

    \hazy{0586} & \DCP{0586} & \BCCR{0586} & \NLD{0586} & \CAP{0586} & \ours{0586} & \clean{0586} \\ \vspace{-1mm}
    \hazy{3146} & \DCP{3146} & \BCCR{3146} & \NLD{3146} & \CAP{3146} & \ours{3146} & \clean{3146} \\ \vspace{-1mm}
    \hazy{4184} & \DCP{4184} & \BCCR{4184} & \NLD{4184} & \CAP{4184} & \ours{4184} & \clean{4184} \\ 
    \hazy{8180} & \DCP{8180} & \BCCR{8180} & \NLD{8180} & \CAP{8180} & \ours{8180} & \clean{8180} \\

    Hazy & DCP \cite{DCP} & BCCR \cite{BCCR} & NLD \cite{NLD} & CAP \cite{CAP} & Ours & Clear \\
    
    \hazy{4561} & \MSCNN{4561} & \AODNet{4561} & \GFN{4561} & \DehazeNet{4561} & \ours{4561} & \clean{4561} \\ \vspace{-1mm}
    \hazy{5576} & \MSCNN{5576} & \AODNet{5576} & \GFN{5576} & \DehazeNet{5576} & \ours{5576} & \clean{5576} \\ \vspace{-1mm}
    \hazy{5920} & \MSCNN{5920} & \AODNet{5920} & \GFN{5920} & \DehazeNet{5920} & \ours{5920} & \clean{5920} \\ \vspace{-1mm}
    \hazy{7471} & \MSCNN{7471} & \AODNet{7471} & \GFN{7471} & \DehazeNet{7471} & \ours{7471} & \clean{7471} \\ \vspace{-1mm}

    Hazy & MSCNN \cite{mscnn} & AOD-Net \cite{aodnet} & GFN \cite{GFN} & Dehaze-Net \cite{dehazenet} & Ours & Clear
    \vspace{-1mm}
    \end{tabular}
    \end{center}
    \caption{Qualitative results on RESIDE's HSTS. Upper half: comparison to prior-based methods; bottom half: comparison to deep-learning-based methods.}
    \label{fig:HSTS_prior_results}
\end{figure*}
\vspace{-1mm}

\begin{figure*}
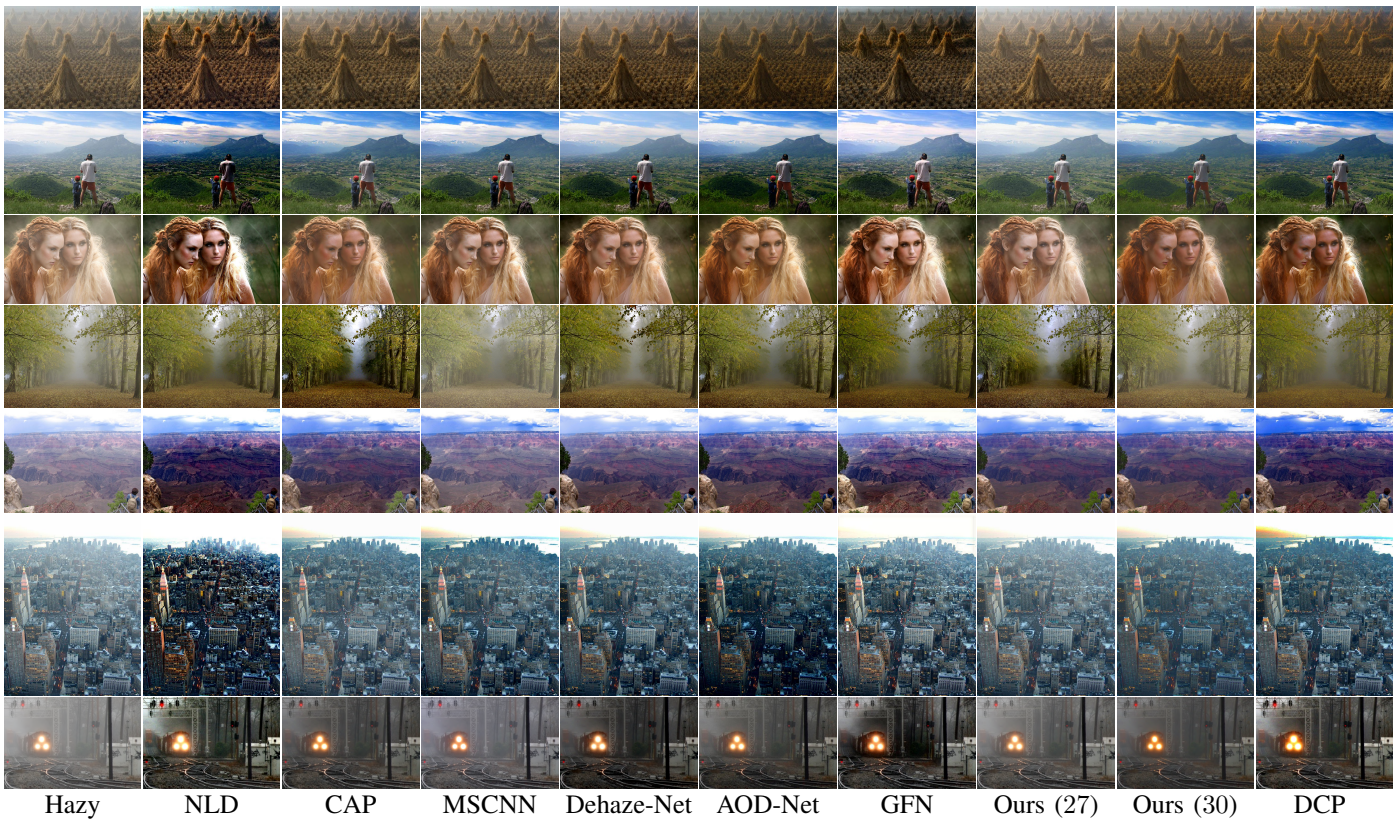

    \begin{center}
    \setlength{\tabcolsep}{0.5pt}
    \begin{tabular}{cccccccccc}

	\vspace{-1mm}

    \hazyq{cones} & \NLDq{cones} & \CAPq{cones} & \MSCNNq{cones} & \DehazeNetq{cones} & \AODNetq{cones} & \GFNq{cones} & \oursq{cones} & \oursp{cones} & \DCPq{cones} \\ \vspace{-1mm}

    %\hazyq{dubai} & \NLDq{dubai} & \CAPq{dubai} & \MSCNNq{dubai} & \DehazeNetq{dubai} & \AODNetq{dubai} & \GFNq{dubai} & \oursq{dubai} & \oursp{dubai} & \DCPq{dubai} \\ \vspace{-1mm}
    
    \hazyq{man} & \NLDq{man} & \CAPq{man} & \MSCNNq{man} & \DehazeNetq{man} & \AODNetq{man} & \GFNq{man} & \oursq{man} & \oursp{man} & \DCPq{man}\\ \vspace{-1mm}
    
    \hazyq{girls} & \NLDq{girls} & \CAPq{girls} & \MSCNNq{girls} & \DehazeNetq{girls} & \AODNetq{girls} & \GFNq{girls} & \oursq{girls} & \oursp{girls} & \DCPq{girls}\\ \vspace{-1mm}

    \hazyq{forest} & \DCPq{forest} & \NLDq{forest} & \CAPq{forest} & \MSCNNq{forest} & \DehazeNetq{forest} & \AODNetq{forest} & \GFNq{forest} & \oursq{forest} & \oursp{forest} \\ \vspace{-1mm}

    \hazyq{hazyDay}  & \NLDq{hazyDay} & \CAPq{hazyDay} & \MSCNNq{hazyDay} & \DehazeNetq{hazyDay} & \AODNetq{hazyDay} & \GFNq{hazyDay} & \oursq{hazyDay} & \oursp{hazyDay} & \DCPq{hazyDay}\\ \vspace{-1mm}

    \hazyq{ny12} & \NLDq{ny12} & \CAPq{ny12} & \MSCNNq{ny12} & \DehazeNetq{ny12} & \AODNetq{ny12} & \GFNq{ny12} & \oursq{ny12} & \oursp{ny12} & \DCPq{ny12} \\ \vspace{-1mm}

    \hazyq{train} & \NLDq{train} & \CAPq{train} & \MSCNNq{train} & \DehazeNetq{train} & \AODNetq{train} & \GFNq{train} & \oursq{train} & \oursp{train} & \DCPq{train} \\ \vspace{-1mm}

    Hazy  & NLD & CAP & MSCNN & Dehaze-Net & AOD-Net & GFN & Ours ($27$) & Ours ($30$) & DCP
	\vspace{-2mm}
    \end{tabular}
    \end{center}
    \caption{Qualitative results of single image dehazing on real-world images. The numbers in parenthesis are the number of epochs spent training our network.}
    \label{fig:real_results}
\end{figure*}

\subsection{Quantitative Evaluation}
We evaluate the performance of our method on the SOTS-indoor, SOTS-outdoor and HSTS test sets. These test sets are created synthetically, therefore featuring both the clean images and their hazy versions. We measure the quality of our solution in terms of the PSNR and SSIM metrics. We obtain the original code and compare our results to the following prior-based approaches: DCP \cite{DCP}\footnote{Implementation in \url{https://github.com/sjtrny/Dark-Channel-Haze-Removal}}, BCCR \cite{BCCR} and NLD \cite{NLD}, and the following data-driven methods: CAP \cite{CAP}, MSCNN \cite{mscnn}, DehazeNet \cite{dehazenet}, AOD-Net \cite{aodnet} and GFN \cite{GFN}. 
In case the dehazed images are out of the range $[0,1]$, we normalize them to $[0,1]$ only if it improves PSNR and SSIM values.

The numeric results\footnote{Our results slightly differ from \cite{reside}. We use the original DCP \cite{DCP}, whereas they use the faster version \cite{fast_DCP} with worse quality.} are given in Table \ref{tbl:results}. We get the highest PSNR and SSIM scores among all other methods in the larger SOTS-outdoor, and the highest SSIM in the smaller HSTS. Our method, represented by a rich neural network and trained to accommodate numerous images, obtains better results compared to prior-based methods. Specifically, compared to DCP, our method strives to approximate the solution of the same energy function, but we stop it before reaching an absolute minimum in order to get further regularization. This is particularly noticed in outdoor images where DCP often over-saturates the sky.

\begin{figure}
    \begin{center}
    \setlength{\tabcolsep}{0.5pt}
    \begin{tabular}{ccc}
	\vspace{-1mm}
    \ITSorig{119_comp} & \ITShazy{119_comp} & \ITStrans{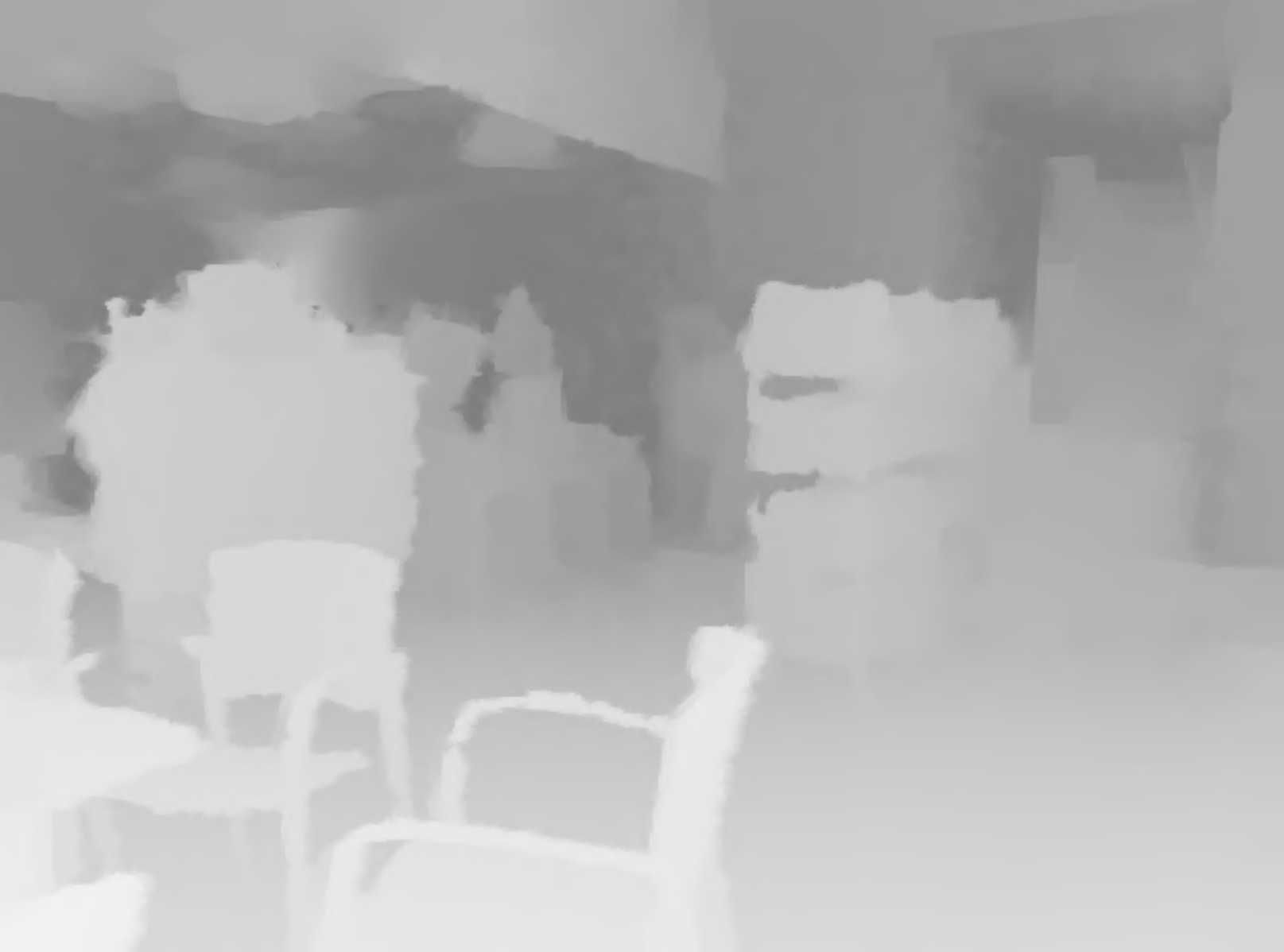} \\ 
    Original & Hazy & Transmission \\ 
    \end{tabular}
    \end{center}
    \caption{Example of a synthetic image and its coarse transmission map from RESIDE's ITS training dataset \cite{reside}.}
    \label{fig:ITS}
\end{figure}

With regard to data-driven approaches, our high score is attributed to the fact that we train on \textit{real-world outdoor} images, whereas competing methods \cite{mscnn,aodnet,GFN} concentrate on \textit{synthetic indoor} images and suffer from a certain domain shift when addressing outdoor data. In addition, the synthetic hazy and clean pairs are created from coarse depth data for which training creates a negative bias towards data-driven approaches. An example of an indoor training image in ITS is given in Figure \ref{fig:ITS}. Notice the rough misplaced edges in the transmission map which later translate to inaccurate hazy images. Indeed, our closest competitor in terms of outdoor results is DehazeNet \cite{dehazenet}. Recall that this method is trained on a large variety of clean image patches of outdoor scenes, making it more robust compared to methods trained on ITS.

We include the results of our method on SOTS-indoor in which it performs favourably, but gets a lower score compared to other data-driven methods and even DCP. This is expected since we train on outdoor images, creating  a tradeoff between indoor and outdoor performance.
As for DCP, it behaves more agreeably on indoor images which coincide better with the haze formation model and do not include sky regions. 

\subsection{Qualitative Results}

We present qualitative results on HSTS in Figure \ref{fig:HSTS_prior_results}. In the top part of Figure \ref{fig:HSTS_prior_results}, it can be seen that our method maintains the true colors of the original image, whereas DCP \cite{DCP}, BCCR \cite{BCCR} and NLD \cite{NLD} tend to produce exaggerated sky regions. Our results are similar to those produced by CAP \cite{CAP}, however slightly closer to the true colors exhibited in the ground truth image. In the bottom half of Figure \ref{fig:HSTS_prior_results} we provide a comparison to deep-learning based methods. In most images we maintain the true contrast and colors, whereas MSCNN \cite{mscnn} and GFN \cite{GFN} provide more contrast-enhanced images. At times, we slightly change the color of the sky, which is to be expected since our method is unsupervised and does not witness the clear images at any stage. In Figure \ref{fig:real_results}, one can see a real-world image comparison of our results with both prior-based and data-driven methods. We display the output of our network after $27$ epochs (the optimal results for the OTS validation set we use) and after $30$ epochs, where the produced images are more similar to DCP (see discussion on sec. \ref{proximity_DCP}). One can see that after $27$ epochs we do not remove all of the haze, perhaps indicating that the outdoor images in RESIDE are less hazy than real-world hazy images. For $30$ epochs, our result is more saturated and of higher contrast.

\subsection{Runtime Comparison}

Apart from improving the overall PSNR and SSIM performance of DCP, we hereby show that we are as efficient as fast implementations of DCP. Our inference procedure consists of two parts: a forward-pass over the trained network to obtain the predicted transmission map (performed in TensorFlow), and reconstruction using Equation \ref{eq:reconstruction} (performed in Numpy). We compare ourselves to a Matlab implementation of soft matting DCP \cite{DCP}, denoted as ``slow-DCP'', and guided image filter DCP \cite{fast_DCP}, denoted as ``fast-DCP''. Note that fast-DCP is an approximation of slow-DCP and though being very efficient, achieves inferior results. Although Matlab is more efficient than Numpy and TensorFlow, we do get the benefit of using the GPU. Thus for fair comparison, we include both GPU and Intel(R) i7-5930k 3.5GHz CPU runtimes of our solution. In Table \ref{tbl:runtimes}, we report the average runtimes (lower is better) over the $500$ images in SOTS-outdoor which feature varying widths of $\sim 500$ pixels. Compared to the explicit optimization of slow-DCP, our feed-forward inference is much faster ($\times 30$ in GPU and $\times 12$ in CPU). We perform on par with fast-DCP (faster for GPU and slower for CPU), but we supply results of much better visual quality which translate to a $\sim 9.5$ dB increase in PSNR. Additional speed-up of our method can be performed by joint estimation of $\bt,\bA$ during training, but we leave this for future work. To conclude, as efficient as the DCP explicit solution may be, it will lack the additional regularization obtained by our approach.

\begin{table} 
 \caption{Average runtime and performance of SOTS-outdoor}
 \label{tbl:runtimes}
 \centering
 \def\arraystretch{1.2}
 \resizebox{0.5\textwidth}{!}{
  \begin{tabular}{lcccc}
    \hline \hline
     & slow-DCP \cite{DCP} & fast-DCP \cite{fast_DCP} & ours-CPU & ours-GPU \\ \hline
    PSNR/SSIM & $17.56/0.822$ & $14.62/0.752$ & $\textbf{24.07}/\textbf{0.933}$ & $\textbf{24.07}/\textbf{0.933}$ \\ \hline
    runtime[sec] & $21.67$ & $1.08$ & $1.71$ & $\textbf{0.67}$ \\ \hline
    \hline 
    %\vspace{-1.3cm}
  \end{tabular}
  }
\end{table}

\section{Discussion}\label{discussion}

\subsection{Proximity to Dark Channel Prior}\label{proximity_DCP}
During training, our network strives to approximate the DCP energy function. Since it optimizes the loss for the entire corpus of images, it may output different results from DCP \cite{DCP}. While DCP operates on one image at a time, our network learns a more ``universal solution'', suited for multiple images. In addition, as the epochs evolve and the loss value decreases, we reach closer and closer to DCP, as can be seen in the three rightmost columns in Figure \ref{fig:real_results}. At earlier epochs the output images still contain a large amount of haze, whereas further on, most of the haze is lifted, but the colors appear more saturated, even non-realistic. We search for a middle-ground where most of the haze is removed and one can see the details, but the colors and contrast remain realistic and physically valid. The benefit of stopping before reaching a deeper optimum of DCP is especially noticeable in sky regions where DCP would output an exaggerated and amped-up version of the sky, whereas we produce a more natural color. In our case this ``sweet-spot'' is reached after $27$ epochs over the training data. Nonetheless, we can keep training for a few more epochs to get more vivid results which may be more pleasant to the human eye.

\subsection{Unsupervised Training Regime}\label{stopping_criterion}
Although our training is completely unsupervised, we do need a stopping criterion since reaching the minimum of the loss function is not always beneficial in terms of the visual and quantitative results. To do so, we evaluate the averge loss value, PSNR and SSIM of a small supervised set of $500$ images from OTS (not part of SOTS-outdoor or HSTS). A typical behaviour of the results is a decrease of the average loss; an increase in performance in PSNR and SSIM; reaching a maximum, and then, a decrease of these criteria. We choose the epoch/model that gave the best performance on the validation set from OTS. The learning parameters are chosen using a similar technique.

\section{Conclusions}\label{conclusions}
We have presented a method of unsupervised training of deep neural networks for the purpose of single image dehazing. Our method relies on the well-known Dark Channel Prior (DCP) \cite{DCP} and manages to improve it considerably. In addition to providing state-of-the-art performance in outdoor scenarios, our method also eliminates the need for synthetic training sets. While our focus here is DCP, we could have incorporated any other successful energy function, using it as our unsupervised loss. Our future research is focused on finding an even better \textit{combination} of energy functions, or incorporating some amount of supervision to benefit from both worlds.

\bibliographystyle{ieeetr}
\bibliography{IEEEabrv,egbib}

\vskip -20pt plus -1fil

\begin{IEEEbiography}[{\includegraphics[width=1in,height=1.25in,clip,keepaspectratio]{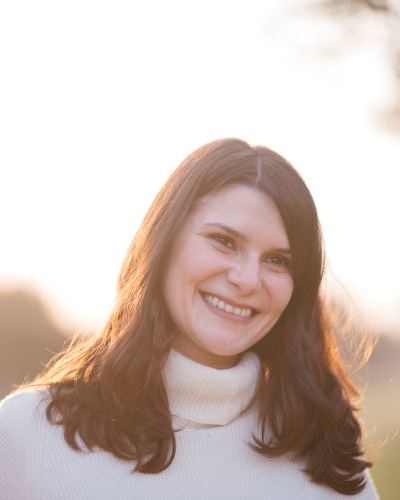}}]{Alona Golts}
received the B.Sc. and M.Sc. degrees from the Department of Electrical Engineering, Technion—Israel Institute of Technology, Haifa, Israel, in 2010 and 2015, respectively. She is
currently pursuing her Ph.D. in the department of Computer Science in the Technion. Her research interests are deep learning, inverse problems and sparse representations.
\end{IEEEbiography}

\vskip -20pt plus -1fil

\begin{IEEEbiography}[{\includegraphics[width=1in,height=1.25in,clip,keepaspectratio]{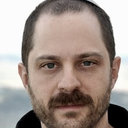}}]{Daniel Freedman}
received the AB in Physics from Princeton University (Magna Cum Laude) in 1993, and his Ph.D. in Engineering Sciences from Harvard University in 2000. From 2000-9, he served as Assistant Professor and Associate Professor in the Computer Science Department at Rensselaer Polytechnic Institute (RPI) (Troy, NY). In 2007, he became a Fulbright Fellow and Visiting Professor of Applied Mathematics and Computer Science at the Weizmann Institute of Science. He then worked in a number of research positions in HP Labs, IBM Research, Microsoft Research, and finally Google Research. In addition to the Fulbright Fellowship, he received the National Science Foundation CAREER Award.
\end{IEEEbiography}

\vskip -20pt plus -1fil

\begin{IEEEbiography}[{\includegraphics[width=1in,height=1.25in,clip,keepaspectratio]{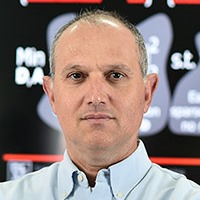}}]{Michael Elad}
received the B.Sc., M.Sc., and
D.Sc. degrees from the Department of Electrical engineering, Technion—Israel Institute of Technology,
Haifa, Israel, in 1986, 1988, and 1997, respectively.
Since 2003, he has been a faculty member in the
Department of Computer Science, Technion—Israel
Institute of Technology, where since 2010, he is a
Full Professor. He works in the field of signal and
image processing, specializing in inverse problems,
and sparse representations. He was the recipient of
numerous teaching awards, the 2008 and 2015 Henri
Taub Prizes for Academic Excellence, and the 2010 Hershel-Rich prize for innovation. He is a SIAM Fellow (2018). Since January 2016, He has been the
Editor-in-Chief for SIAM Journal on Imaging Sciences since January 2016.
\end{IEEEbiography}

\end{document}